\documentclass[11pt,a4paper]{article}
\usepackage{acl2015}
\usepackage{times}
\usepackage{url}
\usepackage{latexsym}
\usepackage{graphicx}
\usepackage{color}
\begin{document}
% You can expand the titlebox if you need extra space
% to show all the authors. Please do not make the titlebox
% smaller than 5cm (the original size); we will check this
% in the camera-ready version and ask you to change it back.
\title{Towards Deep Learning in Hindi NER: An approach to tackle the Labelled Data Scarcity}
\newcommand*{\affaddr}[1]{#1}
\newcommand*{\affmark}[1][]{\textsuperscript{#1}}
\newcommand*{\email}[1]{\texttt{#1}}

\author{%
 Vinayak Athavale\footnotemark[1] \affmark\textsuperscript{[1]},Shreenivas Bharadwaj  \thanks{* indicates these authors contributed equally to this work.} \affmark\textsuperscript{[2]},Monik Pamecha\footnotemark[1] \affmark\textsuperscript{[3]},Ameya Prabhu\affmark\textsuperscript{[1]} and Manish Shrivastava\affmark\textsuperscript{[1]} 
\\ 
{\affmark[1]International Institute of Information Technology, Hyderabad (India)}\\
{\affmark[2]National Institute of Technology, Tiruchirappalli} , 
{\affmark[3]Dwarkadas J. Sanghvi College of Engineering}\\
\affaddr{}
\email{
\{vinayak.athavale,ameya.prabhu\} @research.iiit.ac.in}\\
\email{
\{vshreenivasbharadwaj\} @gmail.com , \{monik.pamecha\}@djsce.edu.in }\\
\email{
\{m.shrivastava\} @iiit.ac.in}%
}

\maketitle

\begin{abstract}
  In this paper we describe an end to end 
  Neural Model for Named Entity Recognition
(NER) which is based on  Bi-Directional RNN-LSTM. Almost all NER systems for Hindi use Language Specific 
features and handcrafted rules with gazetteers. 
Our model is language independent and uses no
domain specific features or any handcrafted
rules. Our models rely on semantic
 information in the form of word vectors
  which are learnt by an unsupervised
learning algorithm on an unannotated corpus.
Our model attained state of the art 
performance in both English and Hindi without the
use of any morphological analysis or without
using gazetteers of any sort.
\end{abstract}

\section{Introduction}

  Named entity recognition (NER) is a very important task in Natural Language Processing. In the NER task, the objective is to find and cluster named entities in text into any desired categories such as person names (PER), organizations (ORG), locations (LOC), time expressions, etc. NER is an important precursor to tasks like Machine Translation, Question Answering , Topic Modelling and Information Extraction among others. Various methods have been used in the past for NER including Hidden Markov models, Conditional Random fields, Feature engineering approaches using Support Vector Machines, Max Entropy classifiers for finally classifying outputs and more recently neural network based approaches.
%Add more text here

Development of an NER system for Indian languages is a comparatively difficult task. Hindi and many other Indian languages provide some inherent difficulties in many NLP related tasks. The structure of the languages contain many complexities like free-word ordering (which affect n-gram based approaches significantly), no capitalization information and its inflectional nature (affecting hand-engineered approaches significantly). Also, in Indian languages there are many word constructions that can be classified as Named Entities (Derivational/Inflectional constructions) etc and these constraints on these constructions vary from language to language hence carefully crafted rules need to be made for each language which is a very time consuming and expensive task. 

Another major problem in Indian languages is the fact that we have scarce availability of annotated data for indian languages. The task is hard for rule-based NLP tools, and the scarcity of labelled data renders many of the statistical approaches like Deep Learning unusable. This complexity in the task is a significant challenge to solve. Can we develop tools which can generalize to other languages(unlike rule based approaches) but still can perform well on this task?

On the other hand, RNN’s and its variants have consistently performed better than other approaches on English NER and many other sequence labelling tasks. We believe RNN would be a very effective method compared to fixed-window approaches as the memory cell takes much larger parts of the sentence into context thus solving the problem of sentences being freely ordered to a large extent. We propose a method to be able to model the NER task using RNN based approaches using the unsupervised data available and achieve good improvements in accuracies over many other models without any hand-engineered features or any rule-based approach. We would learn word-vectors that capture a large number of precise semantic and syntactic word relationships from a large unlabelled corpus and use them to initialize RNNs thus allowing us to leverage the capabilities of RNNs on the currently available data. We believe to the best of our knowledge, that this is the first approach capable of using RNN for NER in Hindi data. We believe learning based approaches like these could generalize to other Indian languages without having to handcraft features or develop dependence on other NLP related tools. Our model uses no language specific features or gazetteers or dictionaries. We use a small amount of supervised training data along with some unannotated corpus for training word embeddings yet we achieve accuracies on par with the state of the art results on the CoNLL 2003 dataset for English  and  achieve 77.48\% accuracy on ICON 2013 NLP tools corpus for Hindi language.  

Our paper is mainly divided into the following sections:
\begin{itemize}
\item In Section 1 we begin with an introduction to the task of NER and briefly describe our approach. 
\item In Section 2, we mention the issues with hindi NER and provide an overview of the  past approaches to NER.
\item In Section 3, we descibe our proposed RNN based approach to the task of NER and the creation of word embeddings for NER which are at the core of our model.
\item In Section 4  We explain our experimental setup, describe the dataset for both Hindi and English and give results and observations of testing on both the datasets.
\item In Section 5 We give our conclusions from the experiments and also describe methods to extend our approach to other languages.
\end{itemize}

\section{Related Work}   

NER task has been extensively studied in the literature. Previous approaches in NER can be roughly classified into Rule based approaches and learning based approaches. Rule based approaches include the system developed by Ralph Grishman in 1995 which used a large dictionary of Named Entities \cite {Grishman:1995}. Another model was built for NER using large lists of names of people, location etc. in 1996\cite {Wakao:1996}. A huge disadvantage of these systems is that a huge list needed to be made and the output for any entity not seen before could not be determined. They lacked in discovering new named entities, not present in the dictionary available and also cases where the word appeared in the dictionary but was not a named entity. This is an even bigger problem for indian languages which would frequently be agglutinative in nature hence creation of dictionaries would be rendered impossible. People either used feature learning based approaches using Hand-crafted features like Capitalization etc. They gave these features to a Machine learning based classifier like  Support Vector Machine (SVM)\cite {Takekuchi:2002}, Naive Bayes (NB) or Maximum Entropy (ME) classifiers. Some posed this problem as a sequence labelling problem terming the context is very important in determining the entities. Then, the handcrafted series were used in sequences using Machine learning methods such as Hidden Markov Models (HMM)\cite {Bikel:1997}, Conditional Random Field (CRF) \cite {Das:2013} and  Decision Trees (DT)\cite {Isozaki:2001}. 

Many attempts have been made to combine the above two approaches to achieve better performance. An example of this is \cite {Srihari:2000} who use a combination of both handcrafted rules along with HMM and ME. More recent approaches for Indian language and Hindi NER are based on CRF’s and include \cite {Das:2013} and \cite {Sharnagat:2013}.  

The recent RNN based approaches for NER include ones by \cite {Lample:2016}. Also, there are many approaches which combine NER with other tasks like \cite {Collobert:2011} (POS Tagging and  NER along with Chunking and SRL tasks)  and \cite {Luo:2015} (combining Entity Linking and NER) which have produced state-of-the-art results on English datasets.

%\includegraphics[scale=0.20]{paper1.png}
%Other approaches are more into domain specific NER like 
\begin{figure*}[t]
\centering
    \includegraphics[width=17cm,height=7cm]{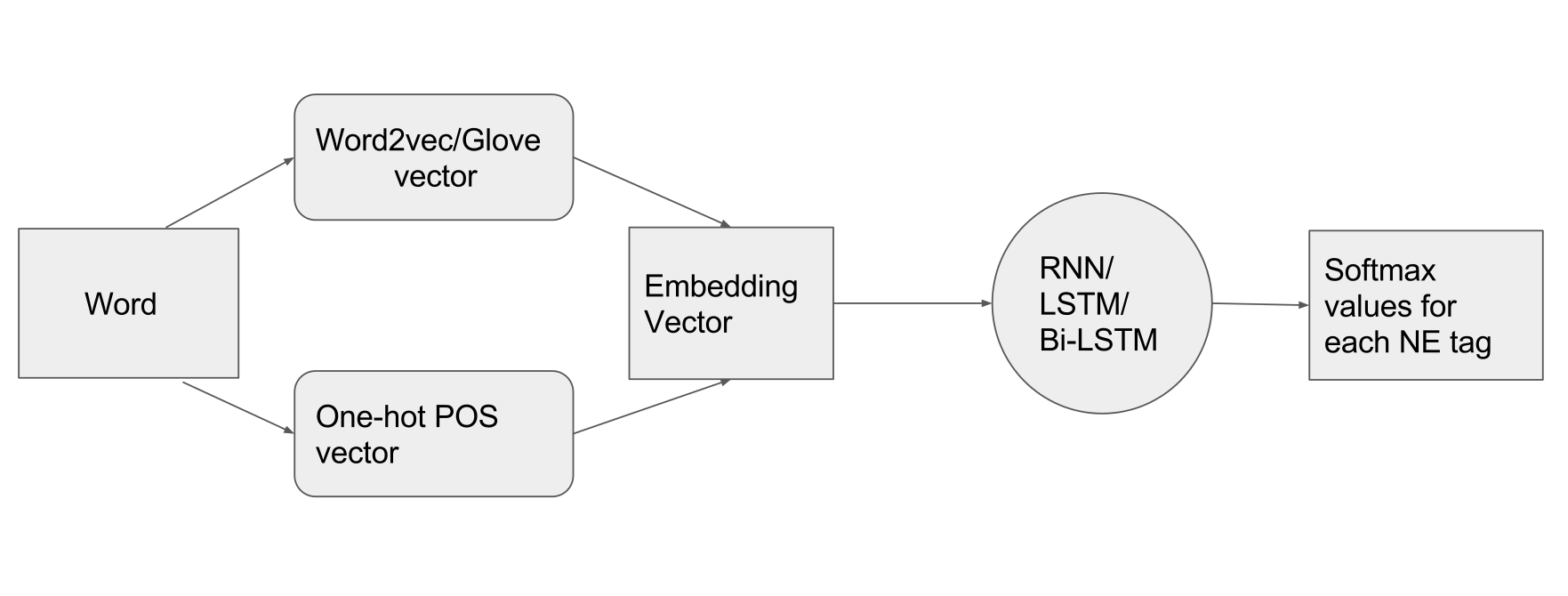}
    \caption{Our pipeline is illustrated above. Every word gets an embedding and a POS tag, which are concatentated to form the word embeddings of the word. It is then passed to recurrent layers and softmax over all classes is the predicted class. }
    \label{fig:methodology}
\end{figure*}

\section{Proposed Approach}

Owing to the recent success in deep learning frameworks, we sought to apply the techniques to Indian language data like Hindi. But, the main challenge in these approaches is to learn inspite of the scarcity of labelled data, one of the core problems of adapting deep-learning approaches to this domain.  

We propose to leverage the vast amount of unlabelled data available in this domain. The recurrent neural networks RNN trained generally have to learn the recurrent layer as well as the embedding layer for every word. The embedding layer usually takes a large amount of data to create good embeddings.  We formulate a two stage methodology to utilize the unlabelled data: 

In the first stage we utilize unlabelled corpora. We learn Skip-gram \cite {Mikolov:2013} based embeddings and GloVe \cite {Pennington:2014} embeddings on those corpora. We use the Wikipedia corpus for Hindi as a source to train these models. By that, we get wordvectors which will be used in the second stage.

In the second stage, as illustrated in Figure 1, we use the deep-learning based models. We initialize their embedding layers with the wordvectors for every word. Then, we train the network end-to-end on the labelled data. As various approaches have proved, a good initialization is crucial to learning good models and train faster \cite {Sutskever:2013}. We apply this approach to use word-vectors to counter the scarcity of labelled data. The idea behind this is that the models would require much lesser data for convergence and would give much better results than when the embeddings are randomly initialized.

To get both previous and subsequent context for making predictions we use Bi-Directional RNN \cite { Schuster :1997}. We know that Vanilla RNN suffers from not being able to model long term dependencies \cite {Bengio:1994} Hence we use the LSTM variant of the RNN \cite {Hochreiter:1997} which helps the RNN model long dependencies better. 

\subsection{Generating Word Embeddings for Hindi}

Word2Vec based approaches use the idea that words which occur in similar context are similar. Thus, they can be clustered together. There are two models introduced by: \cite {Mikolov:2013} CBOW and Skipgram. The latter is shown to perform better on English corpuses for a variety of tasks, hence is more generalizable. Thus, we use the skip-gram based approach. 

Most recent method for generating wordvectors was GloVe, which is similar in nature to that of Skipgram based model. It trains embeddings with local window context using co-occurrence matrices. The GloVe model is trained on the non-zero entries of a global co-occurrence matrix of all words in the corpus. GloVe is shown to be a very effective method, and is used widely thus is shown to be generalizable to multiple tasks in English. 

For English language, we use the pretrained word embeddings using the aforementioned approaches, since they are widely used and pretty effective. The links for downloading the vectors are provided\footnote{Links for download note:webpage not maintained by us
https://github.com/3Top/word2vec-api }. 
However, for Hindi language we train using above mentioned methods(Word2Vec and GloVe) and generate word vectors. We start with One hot encoding for the words and random initializations for their wordvectors and then train them to finally arrive at the word vectors. We use the Hindi text from LTRC IIIT Hyderabad Corpus for training. The data is 385 MB in size and the encoding used is the UTF-8 format (The unsupervised training corpus contains 27 million tokens and 500,000 distinct tokens). The training Hindi word embeddings were trained using a window of context size of 5. The trained model is then used to generate the embeddings for the words in the vocabulary. The data would be released along with the paper at our website along with the wordvectors and their training code\footnote{https://github.com/monikkinom/ner-lstm}. For a comparative study of performance of these methods, we also compare between the Skip-gram based wordvectors and GloVe vectors as embeddings to evaluate their performance on Hindi language. 

\begin{figure}[t]
    \includegraphics[width=\columnwidth]{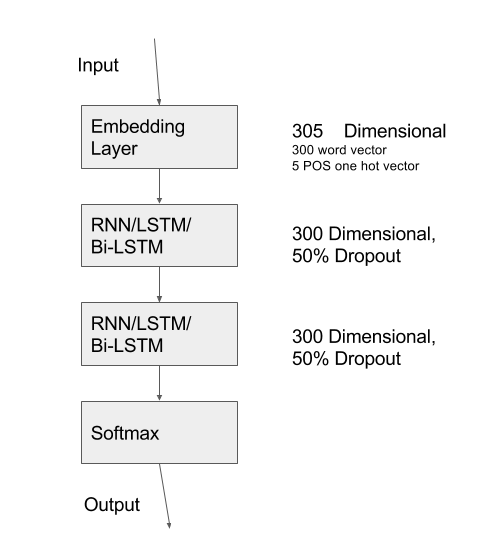}
    \caption{Architecture of the model}
    \label{fig:loss}
\end{figure}

\subsection{Network Architecture}
  
The architecture of the neural networks is described in Figure 2. We trained deep neural networks consisting of either one or two recurrent layers since the labelled dataset was small. In the architecture, we have an embedding layer followed by one or two recurrent layers as specified in the experiments followed by the softmax layer. We experimented with three different kinds of recurrent layers: Vanilla RNN, LSTM and Bi-directional LSTM to test which one would be the most suitable for NER task. For the embedding layer, it is initialized with the concatenation of the wordvector and the one-hot vector indicating its POS Tag. The POS Tagging task is generally considered as a very useful feature for entity recognition, so it was a reliable feature. This hypothesis was validated when the inclusion of POS tags into the embedding improved the accuracy by 3-4\%. 

This setup was trained end-to-end using Adam optimizer \cite {Kingma:2015} and batch size of 128 using dropout layer  with the dropout value of 0.5 after each of the recurrent layers. 
We have used dropout training \cite {Srivastava:2014}  to reduce overfitting in our models and help the model generalise well to the data. The  key  idea  in dropouts is to randomly drop units with their connections from the neural network during training.

\section{Experiments}

We perform extensive experimentation to validate our methodology. We have described the datasets we use and the experimental setup in detail in this section. We then present our results and provide a set of observations made for those results.

\subsection{Datasets}

We test the effectiveness of our approach on ICON 2013 NLP tools contest dataset for Hindi language, along with cross-validating our methodology on the well-established CoNLL 2003 English named entity  recognition  dataset \cite {Sang:2003} .

\subsubsection{ICON 2013 NLP Tools Contest Dataset}

We used the ICON 2013 NLP tools contest dataset to evaluate our models on Hindi. The dataset contains words annotated with part-of-speech (POS) tags and corresponding named entity labels in Shakti Standard Form (SSF) format \cite {Bharti:2009} . The dataset primarily contains 11 entity types: Organization (ORG), Person (PER), Location (LOC), Entertainment, Facilities, Artifact, Living things, Locomotives, Plants, Materials and Diseases. Rest of the corpus was tagged as non-entities (O). The dataset was randomly divided into three splits: Train, Development and Test in the ratios 70\%, 17\% and 13\%. The training set consists of 3,199 sentences comprising 56,801 tokens, development set contains 707 sentences comprising 12,882 tokens and test set contains 571 sentences comprising of 10,396 tokens. We use the F1-measure to evaluate our performance against other approaches.

\subsubsection{CoNLL 2003 Dataset}

We perform extensive experiments on the CoNLL 2003 dataset for Named Entity Recognition. The dataset is primarily a collection of Reuters newswire articles annotated for NER with four entity types: Person (PER), Location(LOC), Organization(ORG), Miscellaneous (MISC) along with non entity elements tagged as (O).  The data is provided with a training set contains 15,000 sentences consisting of approximately 203,000 tokens, along with a development set containing 3466 sentences consisting of around 51,000 tokens and a test set containing 3684 sentences comprising of approximately 46,435 tokens. We use the standard evaluation scripts provided along with the dataset for assessing the performance of our methodology. The scripts use the F1-score to evaluate the performance of models. 

\subsection{Experimental Setup}

We use this architecture for the network because of the constraint on the dataset size caused by scarcity of labelled data. We used a NVIDIA 970 GTX GPU and a  4.00 GHz Intel i7-4790 processor with 64GB RAM to train our models. As the datasets in this domain expand, we would like to scale up our approach to bigger architectures. The results obtained on ICON 2013 NLP Tools dataset are summarized in Table 2. We cross-validated our approach with English language using the CoNLL 2003 dataset. The results are summarized in Table 1, We are able to achieve state-of-the-art accuracies without using additional information like Gazetteers, Chunks along with not using any hand-crafted features which are considered essential for NER task as chunking provides us data about the phrases and Gazetteers provide a list of words which have high likelihood of being a named entity.

\begin{table}[t]
\begin{center}
\resizebox{\linewidth}{!}{%
\begin{tabular}{|l|l|l|l|}
\hline \bf Method & \bf Embed. & \bf Dev & \bf Test \\ \hline
Bi-LSTM			& Random	& 20.04\% & 6.02\% \\
Bi-RNN 		& Skip-300  & 74.30\% & 70.01\% \\
Collobert  & - & - & 89.59\% \\
Luo  (Gaz.)	&   & -- &89.9\% \\
Ours: Bi-LSTM		& Skip-300 & 93.5\% & 89.4\% \\
Ours: Bi-LSTM		& GloVe-300  & 93.99\% & \textbf{90.32\%} \\
Dyer 			& - & - & \textbf{90.94\%} \\
Luo  (Gaz. \& Link)	&   & & \textbf{91.2\%} \\
\hline
\end{tabular}}
\end{center}
\caption{Results on the CoNLL 2003 dataset. We achieve 90.32\% accuracy without using any Gazetter information (Gaz.)  }
\end{table}

\begin{table}[t]
\begin{center}
\resizebox{\linewidth}{!}{%
\begin{tabular}{|l|l|l|l|}
\hline \bf Method & \bf Embed. & \bf Dev & \bf Test \\ \hline

RNN 1l & Skip-300  & 61.88\% & 55.52\% \\
RNN 2l & Skip-300 & 59.77\% & 55.7\% \\
LSTM 1l & Skip-300 & 65.12\% & 61.78\% \\
LSTM 2l & Skip-300 & 61.27\% & 60.1\% \\
Bi-RNN 1l & Skip-300 & 70.52\% & 68.64\% \\
Bi-RNN 2l & Skip-300 & 71.50\% & 68.80\% \\
Bi-LSTM 1l & Skip-300 & 73.16\% & 68.5\% \\
Bi-LSTM 2l & Skip-300 & 74.02\% & 70.9\% \\
Bi-LSTM 1l & GloVe-50 & 74.75\% & 71.97\% \\
Devi et al CRF (Gaz.+Chu.)*  & - &  70.65\% & \textbf{77.44\%} \\
Bi-LSTM 1l & GloVe-300  & \textbf{78.60\%} & \textbf{77.48\%} \\
Das et al CRF (Gaz.)* & - & - & \textbf{79.59\%} \\

\hline
\end{tabular}}
\end{center}
\caption{Results on the ICON NLP Tools 2013 dataset. We achieve 77.48\% accuracy without using any Gazetter information (Gaz.) or Chunking Information (Chu.). }
\end{table}

\begin{table}[t]
\begin{center}
\resizebox{\linewidth}{!}{%
\begin{tabular}{|l|l|l|l|}
\hline \bf Entity Type & \bf Precision & \bf Recall & \bf F1  \\ \hline

ARTIFACT:&    86.04\% &  71.84\% &   78.3\% \\
DISEASE:&      52.5\%  & 80.76\%  &   63.63\% \\
ENTERTAINMENT:& 87.23\% &  84.16\% &   85.66\% \\
FACILITIES:&   56.47\% &  81.35\% &   66.66\% \\
LIVTHINGS:&   55.55\% & 47.61\% &  51.28\% \\
LOCATION:&   26.47\%  & 42.85\% &   32.72\% \\
LOCOMOTIVE:&   60.60\% & 71.42\% &   65.57\% \\
MATERIALS:&  26.31\% & 71.42\% &   38.46\% \\
ORGANIZATION:&   83.33\% &  62.50\% & 71.42\% \\
PERSON:&   61.29\% &   61.29\% &   61.29\% \\
PLANTS:&  50.00\% &  59.99\% &   54.54\% \\
Total: &  75.86\%   & 79.17\% & 77.48\% \\

\hline
\end{tabular}}
\end{center}
\caption{Entity wise Precision, Recall and F1 scores on the ICON NLP Tools 2013 Hindi dataset (Test Set) for glove 300 size Embeddings and Bi-LSTM 1 layer deep model. }
\end{table}

%\afterpage{\clearpage}\
\subsection{Observations}

The neural networks which did not have wordvector based initializations could not perform well on the NER task as predicted. This can be attributed to the scarcity of the data available in the NER task. We also observe that networks consisting of one recurrent layer perform equally good or even  better than networks having two recurrent layers. We believe this would be a validation to our hypothesis that increasing the number of parameters can lead to overfitting.
We could see Significant improvement in performance after using LSTM-RNN instead of Vanilla RNN which can be attributed to the ability of LSTM to model long dependencies.
Also, the bidirectional RNN achieved significant improvement of accuracy over the others suggesting that incorporating context of words around (of both ahead and back) of the word is very useful. 
We provide only 1 layer in our best model to be released along with the paper. \footnote{Code available at: https://github.com/monikkinom/ner-lstm}

\section{Conclusion}

We show that the performance of Deep learning based approaches on the task for entity  recognition can significantly outperform many other approaches involving rule based systems or hand-crafted features. The bidirectional LSTM incorporates features of varied distances providing a bigger context relieving the problem of free-word ordering too. Also, given the scarcity of data, our proposed method effectively leverages LSTM based approaches by incorporating pre-trained word embeddings instead of learning it from data since it could be learnt in an unsupervised learning setting.
We could extend this approach to many Indian Languages as we do not need a very large annotated corpus. When larger labelled datasets are developed, in the new system we would like to explore more deep neural network architectures and try learning the neural networks from scratch.

\footnote{ * indicates that different training/test data is used. Our training data is significantly smaller than the one in consideration (4030 sentence in the original split v/s 3199 sentences in our split.). The original split was unavailable.}

\end{document}